\pdfoutput=1
\documentclass[letterpaper]{article} 
\usepackage{aaai25}  
\usepackage{times}  
\usepackage{helvet}  
\usepackage{courier}  
\usepackage[hyphens]{url}  
\usepackage{graphicx} 
\usepackage{natbib}  
\usepackage{caption} 
\usepackage{algorithm}
\usepackage{algorithmic}
\usepackage{times}  
\usepackage{helvet}  
\usepackage{amsmath}
\usepackage{multirow}
\usepackage{graphicx}
\usepackage{subfigure}
\usepackage{bbding}
\usepackage{makecell}
\usepackage{enumitem}
\usepackage{verbatim}
\usepackage{ragged2e}
\usepackage{booktabs} 
\usepackage{amssymb}
\usepackage{courier}  
\usepackage[hyphens]{url}  
\usepackage{graphicx} 
\usepackage{natbib}  
\usepackage{caption} 
\usepackage{algorithm}
\usepackage{algorithmic}
\usepackage{color}

\usepackage{xcolor}
\usepackage{newfloat}
\usepackage{listings}
\DeclareCaptionStyle{ruled}{labelfont=normalfont,labelsep=colon,strut=off} 
\floatstyle{ruled}
\newfloat{listing}{tb}{lst}{}
\floatname{listing}{Listing}
\title{Knowledge is Power: Harnessing Large Language Models\\ for Enhanced Cognitive Diagnosis}
\author{
    Zhiang Dong,
    Jingyuan Chen\thanks{Corresponding Author.},
    Fei Wu $^{*}$
}
\affiliations{
    Zhejiang University\\

    \{dongza,jingyuanchen,wufei\}@zju.edu.cn
}

\usepackage{bibentry}

\begin{document}

\maketitle

\begin{abstract}
    Cognitive Diagnosis Models (CDMs) are designed to assess students' cognitive states by analyzing their performance across a series of exercises. However, existing CDMs often struggle with diagnosing infrequent students and exercises due to a lack of rich prior knowledge. 
    With the advancement in large language models (LLMs), which possess extensive domain knowledge, their integration into cognitive diagnosis presents a promising opportunity.
    Despite this potential, integrating LLMs with CDMs poses significant challenges. LLMs are not well-suited for capturing the fine-grained collaborative interactions between students and exercises, and the disparity between the semantic space of LLMs and the behavioral space of CDMs hinders effective integration.
    To address these issues, we propose a novel \textbf{K}nowledge-enhanced \textbf{C}ognitive \textbf{D}iagnosis (KCD) framework, which is a model-agnostic framework utilizing LLMs to enhance CDMs and compatible with various CDM architectures. The KCD framework operates in two stages: LLM Diagnosis and Cognitive Level Alignment. In the LLM Diagnosis stage, both students and exercises are diagnosed to achieve comprehensive and detailed modeling. In the Cognitive Level Alignment stage, we bridge the gap between the CDMs' behavioral space and the LLMs' semantic space using contrastive learning and mask-reconstruction approaches.
    Experiments on several real-world datasets demonstrate the effectiveness of our proposed framework.
\end{abstract}

%

  \section{Introduction}

\begin{figure}
    \vspace{-0.5em}
  \centering
  \includegraphics[width=\linewidth]{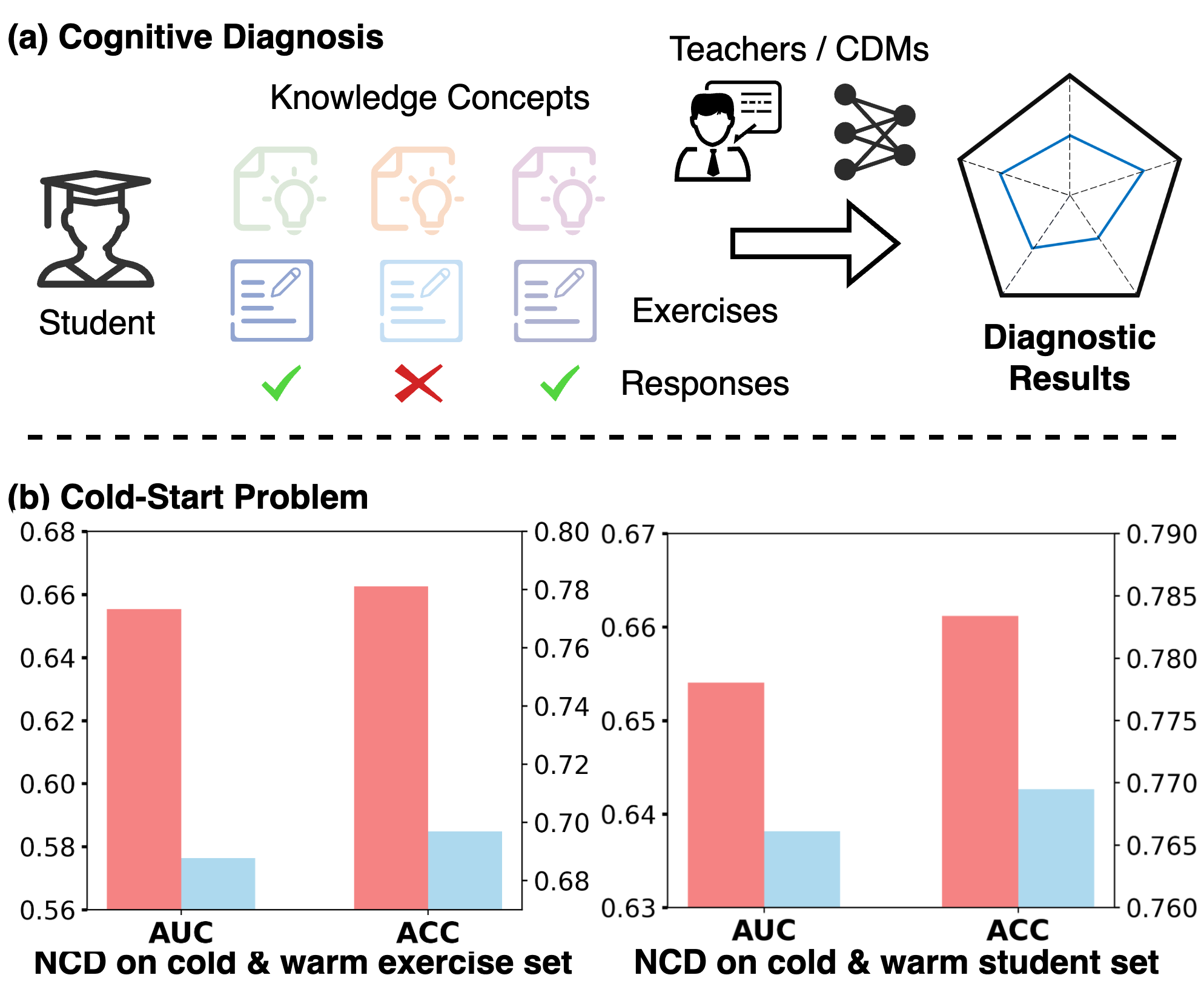}
  \caption{(a) An illustration of cognitive diagnosis.
  (b) Performance of warm and cold scenarios of NCD on PTADisc, exhibiting the limitations in cold scenario.}
    \vspace{-2em}
\label{fig: intro}

\end{figure}

Cognitive diagnosis evaluates a student's learning proficiency through his responses to a series of exercises, as shown in Figure~\ref{fig: intro} (a), which plays a fundamental role in intelligent education systems. 
The outcomes of cognitive diagnosis are crucial for various educational applications, such as educational recommendation~\cite{huang2019exploring}
and computerized adaptive testing~\cite{bi2020quality,zhuang2022robust}.
Consequently, the accuracy and reliability of cognitive diagnosis are essential for enhancing the effectiveness of these educational technologies.

%

Traditional cognitive diagnosis models (CDMs) are primarily grounded in psychometric theories, employing manually designed interaction functions inspired by principles from both psychometrics and educational theory. Examples include DINA~\cite{de2009dina} and MIRT~\cite{reckase200618} models. 
Recent advancements in deep learning have enabled the development of innovative CDMs that leverage neural networks to model complex collaborative information (\textit{i.e.}, student-exercise interactions), thereby improving diagnostic accuracy and adaptability~\cite{wang2020neural, gao2021rcd}. 
However, these existing CDMs face significant challenges in diagnosing infrequent students and exercises, commonly referred to as the cold-start problem. This limitation arises primarily from the lack of prior knowledge within these models, which impairs their adaptability to unfamiliar students and exercises. As depicted in Figure~\ref{fig: intro}(b), experiments conducted on the PTADisc dataset~\cite{hu2023ptadisc} indicate that existing CDMs exhibit poor performance in cold scenarios, thereby undermining overall diagnostic accuracy.

Large language models (LLMs) have seen rapid advancements, showcasing remarkable capabilities in logical reasoning and text comprehension. Their success across various domains highlights the feasibility of this approach~\cite{wang2024user,xu2024eduagent,abbasiantaeb2024let,zhu2024reliable,zhang2024agentcf}. 
The extensive prior knowledge embedded in LLMs presents a promising solution for addressing the limitations of existing CDMs. Specifically, LLMs can leverage their understanding of concepts and relationships between different knowledge domains to provide insights into student learning behaviors and exercise characteristics. By incorporating this extensive prior knowledge, LLMs can simulate the reasoning of experienced human teachers, offering more accurate diagnoses in cold scenarios. Therefore, our objective is to effectively integrate LLMs with CDMs to enhance diagnostic performance. 

However, this integration is non-trivial due to several key factors.
Firstly, LLMs are limited in their ability to model the fine-grained collaborative information crucial for understanding student-exercise interactions, as their input length constraints limit the inclusion of detailed textual information about relationships between exercises, knowledge concepts, and students.
Furthermore, LLMs and CDMs operate in distinct representation spaces: LLMs process text-based data within a semantic space, whereas CDMs analyze student behavior within a behavioral space derived from interactions. 
Successful integration necessitates bridging the gap between these semantic and behavioral spaces.

To address these challenges, we propose a novel Knowledge-enhanced Cognitive Diagnosis (KCD) framework that seamlessly integrates LLMs to enhance existing CDMs, aligning the semantic space of LLMs with the behavioral space of CDMs. 
The proposed KCD comprises two primary modules: \textbf{LLM diagnosis} and \textbf{cognitive level alignment}. The LLM diagnosis module leverages the capabilities of LLMs to simulate experienced human educators in diagnosing students' learning status and the attributes of exercises, thereby enriching the prior knowledge of conventional CDMs. Specifically, during LLM diagnosis, collaborative information regarding students and exercises is gathered via LLMs, followed by an analysis of students' response logs from both educational and psychological perspectives to generate textual diagnoses of students and exercises, revealing their cognitive status and attributes. 
Subsequently, the cognitive level alignment module aligns these textual diagnoses from the semantic space of LLMs with the behavioral representations of CDMs, resulting in more accurate cognitive representations of students.

The contributions of this work are summarized as:

\begin{itemize}
   \item We propose the KCD framework, which is model-agnostic and leverages the combined strengths of LLMs and CDMs to achieve optimal diagnostic results.
   \item We introduce the LLM diagnosis module that combines collaborative information and response logs to generate textual diagnoses of students and exercises.
    \item We introduce the cognitive level alignment module, aligning the textual diagnoses from LLMs with behavioral representations from CDMs.
    \item Experiments on several public datasets with different CDMs demonstrate the effectiveness of our framework. Our code and datasets are available at \url{https://github.com/PlayerDza/KCD}.
\end{itemize}

\section{Related Work}

\subsection{Cognitive Diagnosis}
Cognitive diagnosis, which originated from educational psychology, is a fundamental task in the field of intelligent education. It characterizes students' learning status and knowledge proficiency based on their responses to various questions~\cite{liu2021towards}. Existing cognitive diagnosis methods are mainly divided into two main categories: psychometric theory-based methods~\cite{lord1952theory,de2009dina,reckase200618} and neural network-based methods~\cite{wang2020neural,gao2021rcd,bi2023beta,liu2024inductive,wang2023self}.
Psychometric theory-based methods, such as Item Response Theory (IRT)~\cite{lord1952theory}, Multidimensional IRT (MIRT)~\cite{reckase200618}, and Deterministic Inputs, Noisy And gate model (DINA)~\cite{de2009dina}, are designed to evaluate students' proficiency through latent factors utilizing psychological theories.
Neural network-based methods use deep neural networks to profile students' learning status. NCD~\cite{wang2020neural} first incorporates neural networks into cognitive diagnosis to effectively capture the fine-grained student-exercise relationships. RCD~\cite{gao2021rcd} and RDGT~\cite{yu2024rdgt} employ graph architectures to explore the relationships among exercises, knowledge concepts, and students. Recently, BETA-CD~\cite{bi2023beta} developed a reliable and rapidly adaptable cognitive diagnosis framework for new students through meta-learning. ACD~\cite{wang2024boosting} considered the connection between students' affective states and cognitive states in learning. However, few existing cognitive diagnosis methods take into account prior knowledge, which makes it challenging for them to generate accurate diagnoses. 

\subsection{Large Language Models}
With the rise of Transformer~\cite{vaswani2017attention}, large language models (LLMs) with extensive parameters and vast training data have gradually become mainstream. 
LLMs usually follow a pre-training and fine-tuning approach to accommodate various downstream tasks. They have significantly improved performance in numerous NLP applications, including text summarization~\cite{laskar2022domain,zhang2023summit}, sentiment analysis~\cite{hoang2019aspect,deng2023llms}, translation~\cite{zhang2023prompting,moslem2023adaptive}, and multimodal understanding~\cite{wu2024semantic,huang2024autogeo}.

The advanced comprehension and reasoning capabilities, along with the extensive knowledge repository of LLMs, naturally lead to potential applications in the realm of education.
LLMs can provide researchers with new perspectives by simulating the roles of teachers or students~\cite{wang2024user,li2023adapting,xu2024eduagent,liu2024personality,lin2024e3}, or generating educational resources~\cite{lin2024non,lin2024action,dai2024mpcoder}.
However, less exploration has been made to utilize LLMs for cognitive diagnosis. 
The demonstrated success of LLMs in text summarization tasks and educational contexts indicates LLMs' capability to undertake cognitive diagnostic tasks.

\begin{figure*}[ht]
  \centering
  
  \includegraphics[width=1\linewidth]{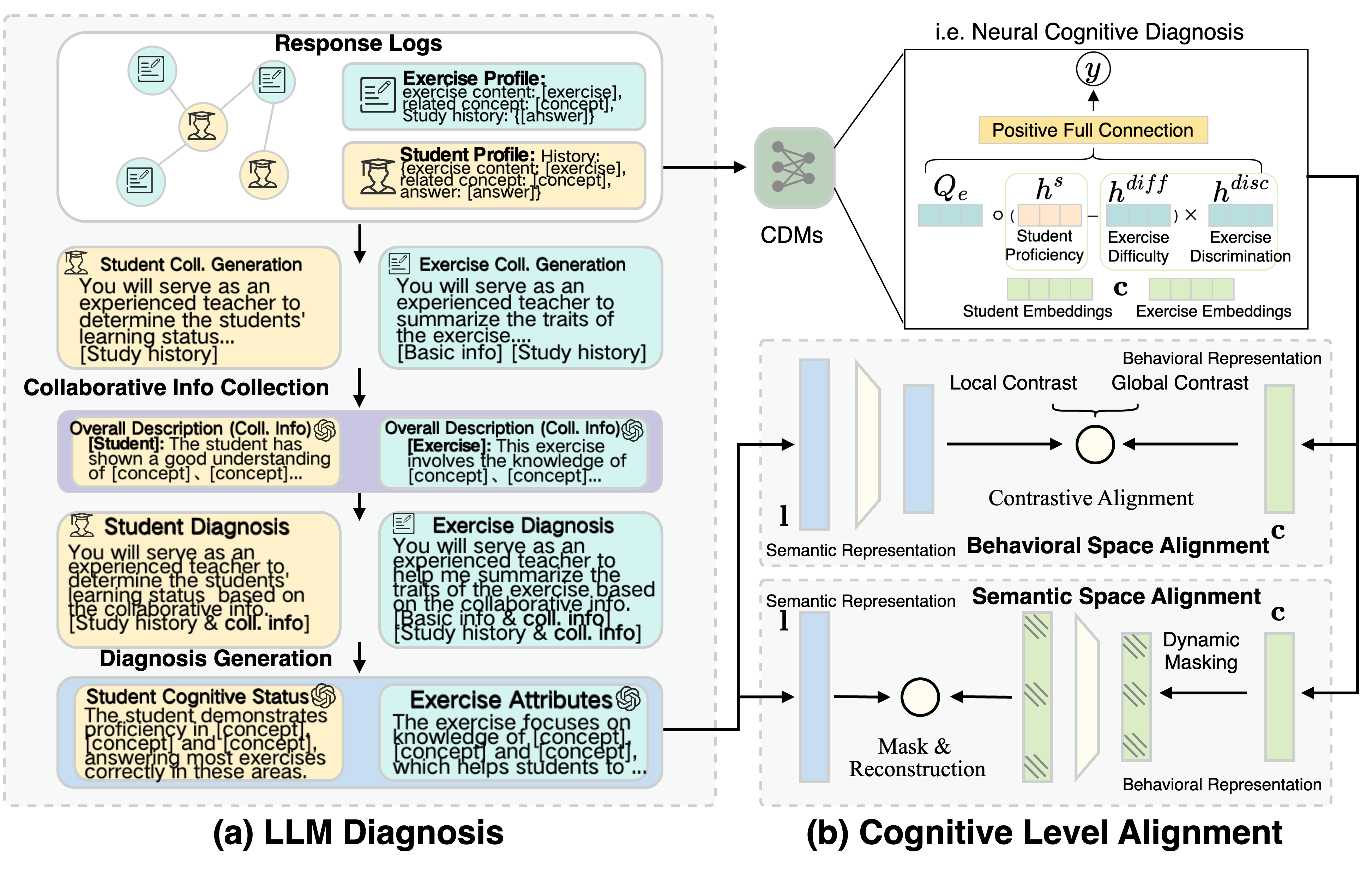}
  \caption{Framework overview. (a) LLM Diagnosis generates diagnoses for students and exercises using LLMs. (b) Cognitive Level Alignment integrates LLMs and CDMs to model students and exercises in both semantic space and behavioral space.}
  \vspace{-1.5em}
\label{fig: method}
\end{figure*}

\section{Methodology}

In this section, we first present the task definition and the general framework. Then, we show the detailed strategies employed within our framework.

\subsection{Task Definition}

Formally, suppose $\mathcal{S} = \{s_1, \cdots, s_{\left|\mathcal{S}\right|}\}$, $\mathcal{E} = \{e_1, \cdots, e_{\left|\mathcal{E}\right|} \}$ and $\mathcal{K} = \{k_1, \cdots, k_{\left|\mathcal{K}\right|} \}$ be the sets of students, exercises and knowledge concepts.
The response logs $\mathcal{R}$ of students are represented as triplets $(s_i, e_j, \mathcal{K}_j, r_{ij}) \in \mathcal{R}$, where $r_{ij}$ indicates whether student $s_i$ correctly answered the exercise $e_j$ and $\mathcal{K}_j$ denotes the knowledge concepts related to $e_j$.
In some datasets, exercise $e$ also includes the text content $t$ as its attributes.
The goal of cognitive diagnosis is to evaluate students' proficiency levels across various knowledge concepts by predicting their performance based on the response logs $\mathcal{R}$.

\subsection{Framework Overview}

The proposed Knowledge-enhanced Cognitive Diagnosis (KCD) framework consists of two main modules: \textbf{LLM diagnosis} and \textbf{cognitive level alignment}, as illustrated in Figure~\ref{fig: method}.
This framework is designed to integrate collaborative information while leveraging the rich prior knowledge of LLMs.
Additionally, it aligns the semantic space of LLMs with the behavioral space of CDMs, thereby combining the strengths of both to optimize diagnostic performance.


The LLM Diagnosis module operates in two stages: collaborative information collection and diagnosis generation. In the first stage, collaborative information is gathered from the response logs. In the second stage, this information, together with the response logs, is utilized to assess students' cognitive statuses and the attributes of exercises. 
The Cognitive Level Alignment module then introduces these LLM-generated diagnoses into conventional CDMs, enhancing the cognitive-level representation of students and exercises. 
This module utilizes two alignment methods, behavioral space alignment and semantic space alignment, to align the textual diagnoses from the semantic space of LLMs and the behavioral space of CDMs.
The framework is model-agnostic, offering flexibility in selecting appropriate CDMs tailored to various educational scenarios, ultimately achieving optimal diagnostic results.

\subsection{LLM Diagnosis}
LLMs can be guided more effectively through carefully crafted natural language instructions, resulting in higher-quality outputs. In this section, we distinguish between two types of input instructions for LLMs: 
system prompts $\mathcal{M}$ and input prompts $\mathcal{P}$. The system prompt $\mathcal{M}$ defines the tasks that LLMs need to perform and specifies the input and output formats, while the input prompt $\mathcal{P}$ consists of specific input data (\textit{i.e.}, students' response logs). 

\subsubsection{Collaborative Information Collection.}
Experienced teachers enhance their diagnoses by utilizing information from other students and exercises. To mimic this capability, we introduce a collaborative information collection stage. This stage aims to extract student collaborative information from a student's performance across all completed exercises and exercise collaborative information from all participating students for a given exercise. 

Specifically, we employ different instruction strategies to diagnose students and exercises. For students, the system prompt $\mathcal{M}_s$ defines the input prompt $\mathcal{P}_s$ format and guides LLMs in generating textual collaborative information. The input prompt $\mathcal{P}_s$ contains the problem content $t$, related knowledge concepts $k$, and the student’s response $r$ for all participated exercises $e$. The input of LLMs is formatted as:
\begin{figure}[!thb]
\centering
\vspace{-1.5em} 
\fcolorbox{black}{gray!6}{%
\parbox{0.9\linewidth}{%
\small
\noindent$\bullet$ \textbf{System Prompt}: You will serve as an experienced teacher to help me determine the student's learning status.
I will provide you with information about exercises that the student has finished, described as STUDY HISTORY, as well as his or her answer of those exercises...
\vspace{0.2em} 
\rule{\linewidth}{0.48pt} 
$\bullet$ \textbf{Input Prompt}:\\
\vspace{-0.1em} 
\hspace{1em} STUDY HISTORY: \\
\vspace{-0.1em}
\hspace{1em} \{content: $t$, concept: $k$, answer: $r$\}; ...\\
}%
}
\label{fig:Fault-Driven-prompt}
\vspace{-1em} 
\end{figure}

Similarly, the input of LLMs for exercises is formatted in the same pattern, where $\mathcal{P}_e$ contains the exercise content $t$, related knowledge concepts $k$, and student responses $r$ for all participating students $s$.
In this way, we can get the collaborative information $\mathcal{I}$ through $\mathcal{I}=\text{LLMs}\left(\mathcal{M}, \mathcal{P}\right)$.

\subsubsection{Diagnosis Generation.}
Once the collaborative information $\mathcal{I}$ for students and exercises has been gathered, the next step is to generate diagnoses of students' cognitive statuses and exercise attributes.
Firstly, we combine the collaborative information $\mathcal{I}$ obtained for each student and exercise with the corresponding response logs to provide more detailed information, formulating input prompt $\mathcal{P}^\prime$.
Then, we adjust the content of system prompt $\mathcal{M}^\prime$ to define the new format of $\mathcal{P}^\prime$ and guide LLMs to generate the corresponding students' cognitive status and exercise attributes. We can get the diagnoses $\mathcal{T}$ through $\mathcal{T}=\text{LLMs}\left(\mathcal{M}^\prime, \mathcal{P}^\prime\right)$. 

\subsection{Cognitive Level Alignment}



By leveraging LLMs, we can generate textual diagnoses of students' cognitive status and exercises' attributes. 
However, LLMs cannot fully comprehend response logs due to constraints on input length, which restrict the inclusion of student-exercise interactions.
Therefore, it is necessary to align these LLM-generated diagnoses with those produced by CDMs at the cognitive level. 
Since LLMs operate within a semantic space while CDMs work within a behavioral space, both need to be mapped to a common space for effective alignment. To achieve this, we propose two alignment methods: behavioral space alignment (KCD-Beh) and semantic space alignment (KCD-Sem).

Before implementing the alignment approach, we obtain the semantic representation of LLMs by encoding their textual diagnoses. Specifically, we utilize the text embedding model~\cite{su2023one}, which has demonstrated significant strength in textual representation, to encode the diagnoses as follows: $\mathbf{L}=\mathbf{E}(\mathcal{T})$,
where $\mathbf{E}(\cdot)$ denotes the text embedding models and $\mathbf{l}\in\mathbf{L}$ denotes the modeling of students and exercises generated by LLMs in semantic space. Meanwhile, we denote the representation embeddings of students and exercises by CDMs as $\mathbf{c}\in\mathbf{C}$ in behavioral space, such as Neural Cognitive Diagnosis (NCD)~\cite{wang2020neural}, as described in Figure~\ref{fig: method}.



\subsubsection{Behavioral Space Alignment.}

Behavioral space alignment involves mapping the LLM-generated models of students and exercises to the behavioral space of CDMs. 
We employ contrastive learning, a widely-used technique for bidirectionally aligning different views~\cite{khosla2020supervised,cui2023generalized}, to align the representations of LLMs and CDMs within the behavioral space.
The intuition behind using contrastive learning is that $\mathbf{c}_i$ and $\mathbf{l}_i$ are most similar to each other within $\mathbf{L}$ since they represent the same student or exercise. 
We apply a multi-layer perceptron (MLP) to map $\mathbf{l}$ from the LLMs’ semantic space to the CDMs’ behavioral space, denoted as $\mathbf{l}^\prime=\mathrm{MLP}(\mathbf{l})$.

Specifically, we conduct contrastive learning from both global and local perspectives. 
Global contrast involves using the entire set $\mathbf{L}$, while local contrast selects a subset $\mathbf{L}^\prime \subset \mathbf{L}$, composed of the $k$ most similar students and exercises for each student and exercise. This subset is obtained by calculating the cosine similarity between each student and exercise with others and selecting the top $k$ most similar instances ($k=20$ in our experiments).
Global contrast captures general features, while local contrast captures fine-grained differences between similar students and exercises.

During training, we use the InfoNCE~\cite{oord2018representation} loss function to calculate both global and local contrast loss values, aiming to maximize the mutual information between $\mathbf{c}$ and $\mathbf{l}$ within the behavioral space, denoted as:
\begin{equation}
    f=-\frac{1}{N} \sum_{i=1}^{N} \log \left(\frac{\exp \left(\frac{x_{i} \cdot y_{i^+}}{\tau}\right)}{\sum_{j=1}^{N} \exp \left(\frac{x_{i} \cdot y_{j^{-}}}{\tau}\right)}\right),
\end{equation}
where $x_{i}$ and $y_{i^+}$ are positive samples, $y_{j^{-}}$ represents the negative sample, $\tau$ denotes the temperature parameter, $N$ denotes the number of samples.
For CDMs, the loss function is denoted as $\mathcal{L}_{cdm}$. For example, the loss function of NCD is cross entropy between output $y$ and ground truth $r$:
\begin{equation}
    \mathcal{L}_{cdm}=-\sum_{i}\left(r_{i} \log y_{i}+\left(1-r_{i}\right) \log \left(1-y_{i}\right)\right).
\end{equation}

The complete loss function is formulated as:
\begin{equation}
    \left\{\begin{matrix}
    \mathcal{L}_{global}=f(\mathbf{c}_i, \mathbf{l}^\prime_i, \mathbf{L}^\prime),   \\
    \mathcal{L}_{local}=f(\mathbf{c}_i, \mathbf{l}^\prime_i, \mathbf{L}^\prime_k),   \\
    \mathcal{L}=\mathcal{L}_{cdm}+\alpha\mathcal{L}_{global}+\beta\mathcal{L}_{local},
    \end{matrix}\right.
\end{equation}
where $f(x_i, x_j, X_k)$ denotes the InfoNCE loss function, $x_i$ and $x_j$ are positive samples, $X_k$ represents the set of negative samples. The term $\mathcal{L}_{cdm}$ denotes the loss function of CDMs. $\mathcal{L}_{global}$ and $\mathcal{L}_{local}$ denote the global contrast loss function and local contrast loss function of behavioral space alignment. $\alpha$ and $\beta$ are hyper-parameters.
By optimizing this loss function $\mathcal{L}$, the CDMs can effectively incorporate the modeling information of students and exercises derived from LLMs, aligning them within the CDMs' behavioral space.

\subsubsection{Semantic Space Alignment.}

In addition to aligning within the behavioral space of CDMs, we also perform alignment in the semantic space of LLMs. Since the semantic space encapsulates rich features of students and exercises, inspired by masked autoencoders (MAE)~\cite{he2022masked}, we utilize a mask-reconstruction strategy to align the representations of the two models. 
Initially, a multi-layer perceptron (MLP) maps $\mathbf{c}$ from the CDMs' behavioral space to the LLMs' semantic space, denoted as $\mathbf{c}^\prime=\mathrm{MLP}(\mathbf{c})$. We then apply a dynamic masking strategy that varies the mask ratio based on the frequency of occurrence of students and exercises. For frequently occurring instances, we increase the mask ratio to extract more semantic information from LLMs, while for less frequent instances, we reduce the mask ratio to minimize the introduction of noise. This process is represented as:
\begin{equation}
    \widehat{\mathbf{c}_i} =\mathrm{MASK}(\mathbf{c}_i, ratio_i),
\end{equation}
where $\widehat{\mathbf{c}_i}$ represents the masked embeddings of student $i$, and $ratio_i$ denotes the mask ratio applied.

During training, the InfoNCE loss function is employed again to calculate the reconstruction loss $\mathcal{L}_{recon}$, which maximizes mutual information between $\mathbf{c}$ and $\mathbf{l}$, thereby aiding the accurate reconstruction of $\mathbf{c}$. The overall loss function is defined as:

\begin{equation}
    \left\{\begin{matrix}
    \mathcal{L}_{recon}=f(\widehat{\mathbf{c}_i}^\prime, \mathbf{l}, \mathbf{L}), \\
    \mathcal{L}=\mathcal{L}_{cdm}+\lambda\mathcal{L}_{recon},
    \end{matrix}\right.
\end{equation}

The hyperparameter $\lambda$ controls the influence of the reconstruction loss on the overall optimization. By optimizing this combined loss function $\mathcal{L}$, the model can produce more accurate and robust representations by reconstructing the masked inputs, thereby aligning the representations from both CDMs and LLMs within the semantic space. This alignment enhances the ability of CDMs to incorporate the rich semantic information provided by LLMs, leading to improved diagnostic accuracy and robustness.


\begin{table}[tbp]

\scalebox{0.85}{
\begin{tabular}{l|cccc}
\toprule
             & Python   & Linux     & Database  & Literature    \\ 
\midrule
\#Students  & 22,953           & 1,253     & 11,891  &  2,264       \\ 
\#Exercises  & 11,807         & 1,335     & 3,106    &  885     \\
\#Concepts  & 713         & 221     & 313    &  31     \\
\#Response Logs  &   170,844         & 34,758       & 122,642    &  30,273       \\ 
\#Logs per Student & 7.44           & 27.74       & 10.38    &   13.37     \\ 
\#Sparsity (\%)  & 0.063           & 2.078       & 0.334     &   1.511    \\ 
\bottomrule
\end{tabular}}
\caption{Statistics of datasets.}
\label{tab: dataset}
\vspace{-1em}
\end{table}
\section{Experiments}
\label{sec: exp}

In this section, we conduct experiments to answer the following research questions:
\begin{itemize}
\item \textbf{RQ1}: Can the proposed model effectively improve the performance of the original CDMs?  
\item \textbf{RQ2}: What is the impact of each component within the proposed method? 
\item \textbf{RQ3}: How does the proposed model perform on cold-start scenarios? 
\item \textbf{RQ4}: How effective is the alignment of semantic and behavioral space embeddings during the cognitive level alignment process?
\end{itemize}

\subsection{Experimental Settings}

\subsubsection{Datasets}

\begin{table*}[tbp]
    \centering

    \scalebox{0.85}{
    \begin{tabular}{lcccccccccccc}
        \toprule
        \multirow{2}{*}{Methods} & \multicolumn{3}{c}{Python} & \multicolumn{3}{c}{Linux} & \multicolumn{3}{c}{Database} & \multicolumn{3}{c}{Literature}\\
        \cmidrule(r){2-4} \cmidrule(r){5-7} \cmidrule(r){8-10} \cmidrule(r){11-13}
        
         & AUC $\uparrow$  & ACC $\uparrow$ & RMSE $\downarrow$ & AUC $\uparrow$ & ACC $\uparrow$ & RMSE $\downarrow$ & AUC $\uparrow$  & ACC $\uparrow$ & RMSE $\downarrow$ & AUC $\uparrow$ & ACC $\uparrow$ & RMSE $\downarrow$\\ 
         \midrule
        IRT & 0.6338 & 0.7749 & 0.4031 & 0.8146 & 0.7874 & 0.3943  & 0.7312 & 0.7989 & 0.3948  & 0.8086 & 0.7818 & 0.3866\\ 
        IRT-Beh & 0.6567 & 0.7966 & 0.3914 & 0.8325 & 0.8048 & 0.3758  & 0.7522 & 0.8022 & 0.3752 & 0.8221 & 0.8006 & 0.3659\\
        IRT-Sem & 0.6621 & 0.7935 & 0.3851 & 0.8286 & 0.7983 & 0.3786  & 0.7435 & 0.8098 & 0.3807 & 0.8197 & 0.7924 & 0.3686\\
        \midrule
        MIRT & 0.6434 & 0.7693 & 0.4415 & 0.8183 & 0.7899 & 0.4056  & 0.7220 & 0.7973 & 0.4217 & 0.8289 & 0.8002 & 0.4083\\ 
        MIRT-Beh & 0.6873 & 0.8049 & 0.4081 & 0.8329 & 0.8067 & 0.3841  & 0.7616 & 0.8178 & 0.4073 & 0.8583 & 0.8225 & 0.3679\\
        MIRT-Sem & 0.6647 & 0.7875 & 0.4186 & 0.8315 & 0.8011 & 0.3874  &0.7443 & 0.8092 & 0.4065 & 0.8465 & 0.8175 & 0.3822\\
        \midrule
        
        DINA & 0.6001 & 0.5521 & 0.4962 & 0.6791 & 0.5469 & 0.4964  & 0.6581 & 0.5981 & 0.4716 & 0.7021 & 0.6162 & 0.4735\\ 
        DINA-Beh & 0.6476 & 0.6213 & 0.4355 & 0.7239 & 0.6094 & 0.4437  & 0.6927 & 0.6713 & 0.4291 & 0.7449 & 0.6992 & 0.4361\\
        DINA-Sem & 0.6354 & 0.6086 & 0.4487 & 0.7032 & 0.6164 & 0.4563  & 0.6792 & 0.6596 & 0.4378 & 0.7263 & 0.6697 & 0.4574\\
        
        \midrule
        NCD & 0.6522 & 0.7758 & 0.4027 & 0.8256 & 0.7759 & 0.3926   & 0.7375 & 0.7932 & 0.3953 & 0.8449 & 0.7805 & 0.3896\\ 
        NCD-Beh & 0.6804 & 0.8007 & 0.3866  &  0.8422 & 0.7928 & 0.3781 & 0.7552 & 0.8205 & 0.3715 & 0.8691 & 0.8183 & 0.3645\\
        NCD-Sem & 0.6687 & 0.7940 & 0.3892 &  0.8460 & 0.7963 & 0.3764 &0.7509 & 0.8170 & 0.3776 & 0.8615 & 0.8076 & 0.3674\\
        \midrule
        RCD & 0.6781 & 0.7767 & 0.3901 & 0.8557 & 0.8086 & 0.3865  & 0.7583 & 0.7948 & 0.3897 & 0.8494 & 0.7879 & 0.3809\\ 
        RCD-Beh & 0.6980 & 0.7945 & 0.3776 & 0.8736 & 0.8292 & 0.3625  & 0.7872 & 0.8194 & 0.3737 & 0.8640 & 0.8151 & 0.3667\\
        RCD-Sem & 0.6904 & 0.7902 & 0.3802 & 0.8715 & 0.8271 & 0.3643  & 0.7849 & 0.8132 & 0.3743 & 0.8598 & 0.8132 & 0.3671\\
        \midrule
        SCD & 0.6815 & 0.7792 & 0.3882 & 0.8594 & 0.8113 & 0.3806  & 0.7598 & 0.7973 & 0.3824 & 0.8537 & 0.7902 & 0.3781\\ 
        SCD-Beh & 0.7023 & 0.7982 & 0.3746 & 0.8751 & 0.8319 & 0.3584  & 0.7934 & 0.8229 & 0.3683 & 0.8695 & 0.8213 & 0.3609\\
        SCD-Sem & 0.6957 & 0.7945 & 0.3779 & 0.8721 & 0.8296 & 0.3608  & 0.7890 & 0.8203 & 0.3697 & 0.8681 & 0.8196 & 0.3624\\
        \midrule
        ACD & 0.6738 & 0.7932 & 0.4007 & 0.8374 & 0.7573 & 0.4079  & 0.7578 & 0.8137 & 0.3786 & 0.8517 & 0.7924 & 0.3765\\ 
        ACD-Beh & 0.7056 & 0.8053 & 0.3839 & 0.8551 & 0.8003 & 0.3734  & 0.7731 & 0.8324 & 0.3626 & 0.8725 & 0.8119 & 0.3602\\
        ACD-Sem & 0.7035 & 0.8004 & 0.3782 & 0.8513 & 0.7737 & 0.3856  &0.7774 & 0.8253 & 0.3678 & 0.8674 & 0.8092 & 0.3629\\
        
        \bottomrule
    \end{tabular}}
    \caption{Performance comparison with baseline methods. The improvements are statistically significant where $p<0.05$.
    }
    \label{tab:performance}
    \vspace{-1em}

\end{table*}

In our experiments, we utilize four courses, Python Programming (Python), Linux System (Linux), Database Technology and Application (Database), and Literature and History (Literature), from a publicly available dataset PTADisc~\cite{hu2023ptadisc}, which comes from real-world students' responses in the educational website PTA\footnote{\url{https://pintia.cn/}} and contains textual information of exercises and knowledge concepts. 
The statistics of the datasets are presented in Table~\ref{tab: dataset}.
The datasets are divided into training, validation, and testing sets, with a ratio of 8:1:1.

\subsubsection{Evaluation Metrics}

Following previous works, we evaluate the students' cognitive status by predicting the performance of students on the testing set, as the cognitive status can not be directly observed. We adopt commonly used metrics, namely the Area Under a ROC Curve (AUC), the Prediction Accuracy (ACC), and the Root Mean Square Error (RMSE), to validate the effectiveness of the CDMs.
For all the metrics, $\uparrow$ represents that a greater value is better, while $\downarrow$ represents the opposite.

\subsubsection{Baseline Methods}

To validate the effectiveness of the proposed method, we conduct experiments on several representative CDMs, including IRT~\cite{lord1952theory}, MIRT~\cite{reckase200618}, DINA~\cite{de2009dina}, NCD~\cite{wang2020neural}, RCD~\cite{gao2021rcd}, SCD~\cite{wang2023self} and ACD~\cite{wang2024boosting}.

\subsubsection{Implementation Details}

We utilize PyTorch to implement both the baseline methods and our proposed KCD framework. 
For the baseline models, We use the default hyper-parameters as stated in their papers and for KCD, we use the same hyper-parameter settings, such as training epoch, learning rate, and batch size.
We employ ChatGPT to represent LLMs (specifically, gpt-3.5-turbo-16k) and text-embedding-ada002 as the text embedding model. All the experiments are conducted on a GeForce RTX 3090 GPU.
We train the model on train set and at the end of each epoch, we evaluate the model on the validation set.
The hyper-parameter $\alpha$, $\beta$, and $\lambda$ was set to $0.04$, $0.015$, and $0.2$.
Since our dataset does not include affect labels, we utilize the unsupervised contrastive ACD model and employ NCD as the basic cognitive diagnosis module.
The behavioral space alignment approach is denoted as `-Beh' and the semantic space alignment approach is denoted as `-Sem'.

\begin{figure}[t]
  \centering
  
  \includegraphics[width=1.02\linewidth]{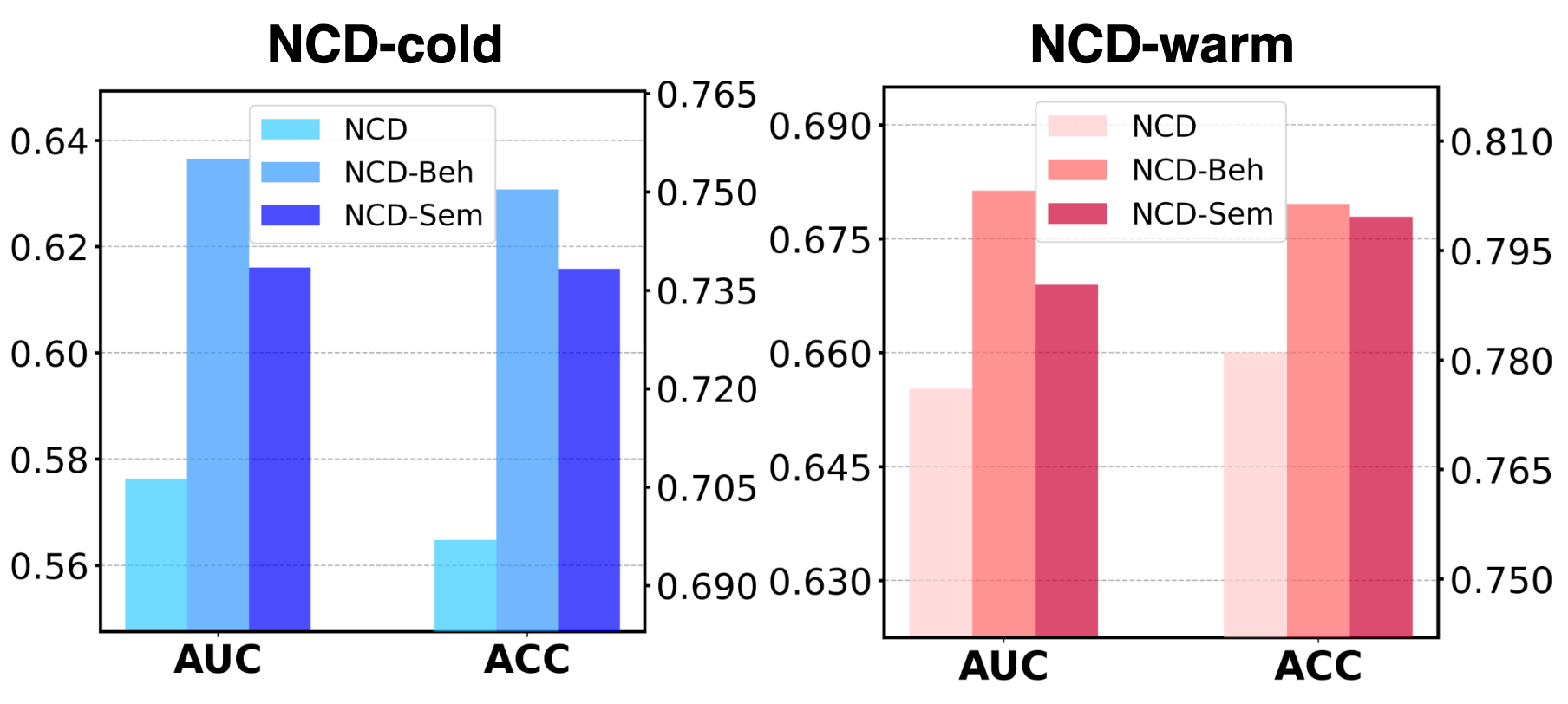}
  \caption{Performance comparison in cold (blue) and warm (red) scenarios on Python dataset.}
  \vspace{-2em}
\label{fig: experiment1}
\end{figure}

\subsection{Performance Comparison (RQ1)}
To demonstrate the effectiveness of our proposed method in improving cognitive diagnosis, we implement the framework on seven cognitive diagnosis models, and the results are shown in Table~\ref{tab:performance}. 
Additionally, we compared the performance of NCD in warm and cold scenarios, with the results illustrated in Figure~\ref{fig: experiment1}. Here we define the cold scenario as less than $3$ interactions in the training set for exercises and define the warm scenario as more than $10$ interactions in the training set for exercises. Following this definition, we divide the testing set into cold and warm subsets.
We have the following observations from the results: 

\begin{itemize}[leftmargin=*]
    \item[1)]  
    Both KCD-Beh and KCD-Sem achieve significant improvements compared to the basic CDMs.
    This indicates that our proposed framework is widely applicable to various CDMs, and both alignment methods can effectively align the behavioral space of CDMs and the semantic space of LLMs.
    In most models, the behavioral space alignment approach performs better, indicating that aligning in the behavioral space of CDMs can better align information from the semantic space of LLMs.
    \item[2)] Compared to basic CDMs, our proposed methods demonstrate improvements in both cold and warm scenarios, especially in cold scenarios. This indicates that our approach of introducing LLMs as knowledge enhancement effectively alleviates the cold-start issue.
\end{itemize}

\begin{table}[t]
\centering

\vspace{-0.5em}

\scalebox{0.9}{
\begin{tabular}{lcccc}
\toprule
Condition & Method &  AUC  & ACC & RMSE \\ \midrule
& NCD & 0.6522 &  0.7758 & 0.4027 \\ 
& NCD-Beh & 0.6804 & 0.8007 & 0.3866 \\
& NCD-Sem & 0.6687 & 0.7940 & 0.3892 \\
\midrule
\multirow{2}{*}{w/o Coll. Info} &NCD-Beh & 0.6765 & 0.7967 & 0.3895 \\
&NCD-Sem & 0.6637 & 0.7884 & 0.3926  \\
\midrule
w/o Local Con. &NCD-Beh & 0.6726  & 0.7872  &  0.3901\\  

w/o Global Con. &NCD-Beh & 0.6748  & 0.7885 & 0.3938 \\
\midrule
w/o Dym. Mask&NCD-Sem & 0.6619  & 0.7876 & 0.3941\\

\bottomrule

\end{tabular}}
\caption{Ablation study on Python dataset. `Coll. Info' denotes collaborative information, while `Local Con.' and `Global Con.' represent local contrast and global contrast. `Dym. Mask' denotes the dynamic masking strategy.} 
\label{tab:ablation}
\vspace{-0.5em}
\end{table}

\subsection{Ablation Study (RQ2)}

To validate the effectiveness of different components of our proposed method, we conduct ablation experiments to verify several components utilized in LLM Diagnosis and Cognitive Level alignment, including the usage of collaborative information (denoted as `Coll. Info'), the local contrast and global contrast (denoted as `Local Con.' and `Global Con.'), and the dynamic masking strategy (denoted as `Dym. Mask').

Table~\ref{tab:ablation} demonstrates the results of the ablation study on Python dataset, comparing the model performance after removing specific components (denoted as `w/o'). `w/o Coll. Info' represents replacing collaborative information in the process of diagnosis generation and `w/o Dym. Mask' represents replacing dynamic masking strategy with a constant mask ratio.
Experimental results show that removing these components individually leads to a decline in the model's performance. This indicates that these components are crucial for the model's performance.

\begin{figure}[t]
  \centering
  
  \includegraphics[width=1\linewidth]{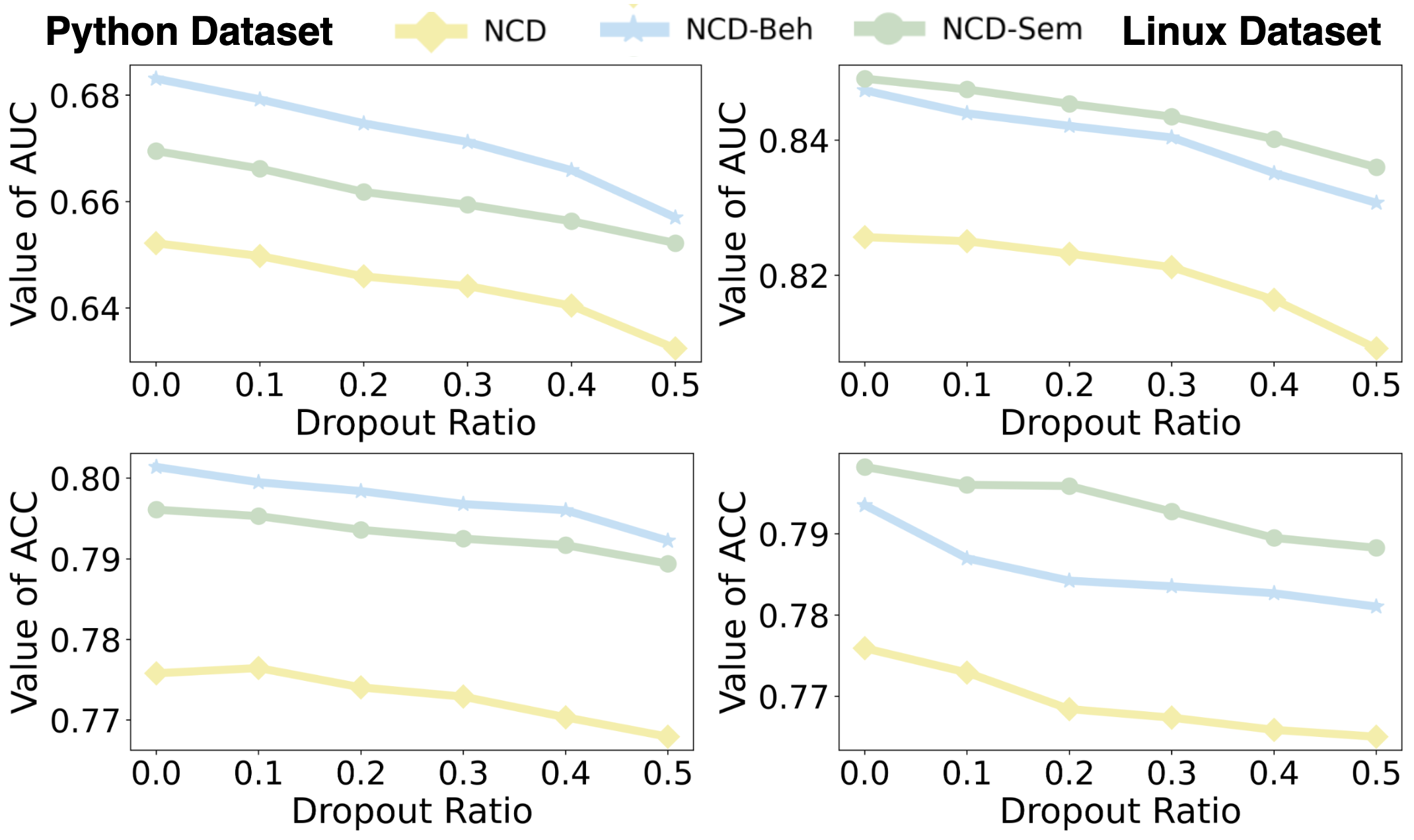}
  \caption{Performance on different dropout ratios.}
  
\label{fig: drop}
\end{figure}
\subsection{Performance on Cold-Start Scenarios (RQ3)}

we conduct additional experiments on sub-datasets with varying degrees of sparsity. Specifically, we apply random dropout to the training sets of the Python and Linux datasets at ratios of $10\%$, $20\%$, $30\%$, $40\%$, and $50\%$.

Figure~\ref{fig: drop} shows the results of the experiments on different dropout ratios. It is obvious that as the dropout ratio increases, both AUC and ACC decrease. This is because the training set becomes more sparse, approaching a cold-start scenario. 
Additionally, compared to ACC, AUC experiences a greater decline, which might be due to the different calculation methods of the two metrics. 

\begin{figure}[t]
  \centering
  \vspace{-1em}
  \includegraphics[width=1\linewidth]{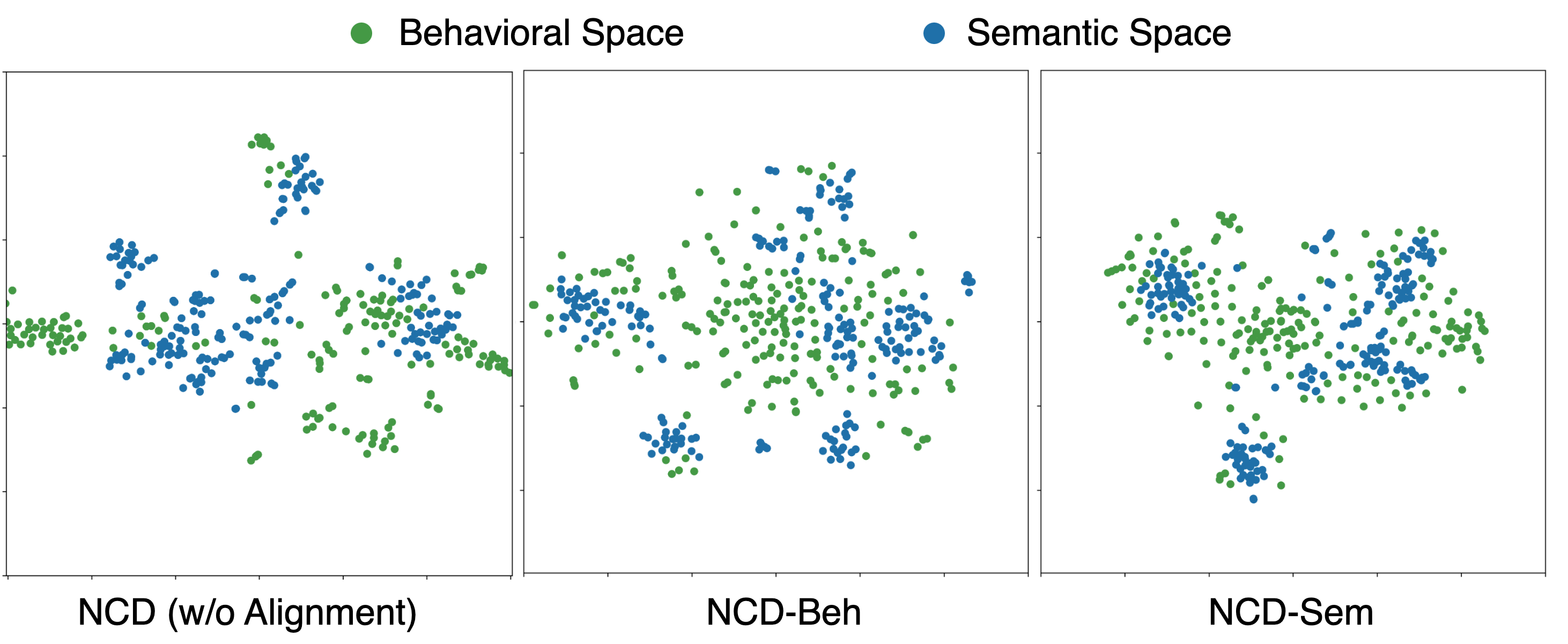}
  \caption{The t-SNE visualization of student embeddings on Literature dataset.}
  \vspace{-2em}
\label{fig: experiment2}
\end{figure}
\subsection{Visualization of Semantic and Behavioral Embeddings (RQ4)}

To validate the effectiveness of the two alignment processes, we utilize t-SNE~\cite{van2008visualizing} to visualize the distribution of features in LLMs semantic space and CDMs behavioral space. We randomly select 200 example students and map their behavioral embeddings and semantic embeddings to 2-dimensional space. NCD (w/o Alignment) represents the original CDMs without alignment.

Figure~\ref{fig: experiment2} demonstrates the integration of semantic and behavioral embeddings of NCD-Beh and NCD-Sem, with their distributions closely merged compared to original CDMs. This proves the effectiveness of the two alignment methods we proposed.

\begin{figure}[t]
  \centering
  
  \includegraphics[width=1\linewidth]{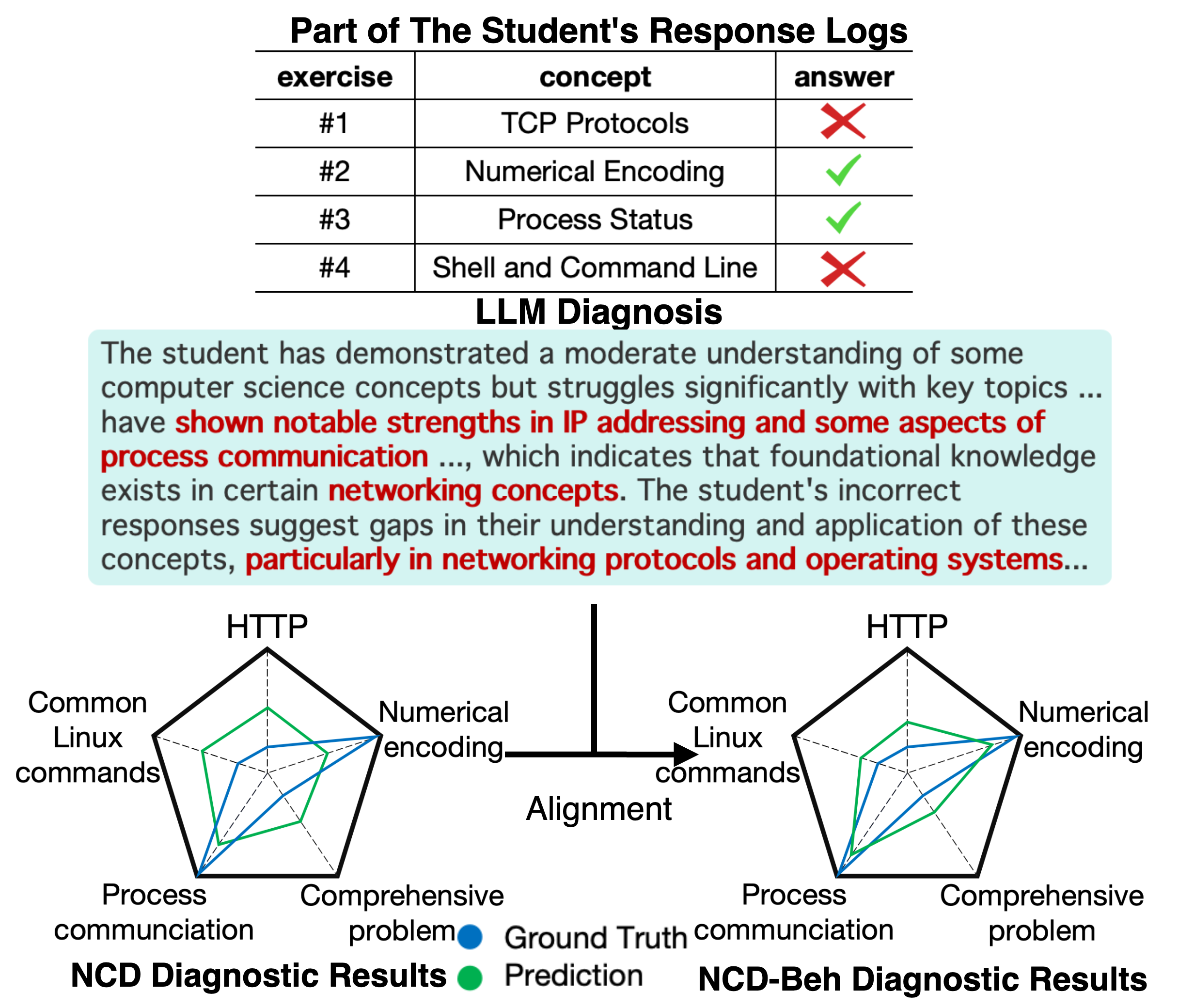}
  \caption{The case study of a student on multiple knowledge concepts on Linux dataset.}
  \vspace{-2em}
\label{fig: case}
\end{figure}

\subsection{Case Study}

To more intuitively demonstrate the improvements our proposed methods bring to CDMs, we selected a diagnosis for a specific student in the Linux dataset and compared the prediction results of NCD with the diagnosis results of NCD-Beh.
As illustrated in Figure~\ref{fig: case}, we randomly choose a student, and list his mastery of some knowledge concepts predicted by NCD and our proposed NCD-Beh.
This student correctly answered the exercises related to `numerical encoding' and `process communication', showing mastery of these concepts. He answered other exercises incorrectly, indicating a lack of familiarity with the remaining knowledge concepts.
From the LLM's diagnostic results, it can be observed that the LLM captured similar question-answer information from the training set and made corresponding inferences. This played an important role in NCD-Beh's more accurate prediction of the student's mastery level.

\section{Conclusion}

In this work, we propose a model-agnostic framework KCD that can efficiently employ LLMs to enhance the knowledge of conventional CDMs.
By utilizing LLM diagnosis and cognitive level alignment, the framework can leverage the rich knowledge of LLMs and align the semantic space of LLMs and the behavioral space of CDMs to achieve optimal diagnostic results.
Several experiments on four real-world datasets for cognitive diagnosis demonstrate the superiority of our proposed framework, surpassing all the baseline CDMs.

\section{Acknowledgments}
This research was partially supported by grants from the National Natural Science
Foundation of China (No.62037001, No.62307032), Shanghai Rising-Star Program
(23QA1409000), the Starry Night Science Fund at Shanghai Institute for Advanced Study (SN-ZJU-SIAS-0010), and the "Pioneer" and "Leading Goose" R\&D Program of Zhejiang under Grant No. 2025C02022.

\bibliography{aaai25}

\end{document}